\documentclass[letterpaper, 10 pt, conference]{ieeeconf}  

\IEEEoverridecommandlockouts                              

\usepackage{amssymb}
\usepackage{graphicx}
\newtheorem{definition}{Definition} 
\usepackage{booktabs} 
\usepackage{multirow}
\usepackage{makecell}
\usepackage{threeparttable}
\usepackage[misc]{ifsym}

\usepackage{amsmath,amsfonts}
\usepackage{algorithmic}
\usepackage{algorithm}
\usepackage{array}
\usepackage[caption=false,font=footnotesize,labelfont=rm,textfont=rm]{subfig}
\usepackage{textcomp}
\usepackage{stfloats}
\usepackage{url}
\usepackage{verbatim}
\usepackage{caption}
\usepackage{cite}
\usepackage{epigraph}
\usepackage{array}
\usepackage[dvipsnames]{xcolor}         

\usepackage[bookmarks=true,breaklinks=true,colorlinks,linkcolor=RedViolet,citecolor=blue,urlcolor=BlueViolet]{hyperref}

\newcommand{\myAPP}{SFCF}

\author{Leizhen Zhang$^{1}$, Lusi Li$^{2}$, Di Wu$^{3}$, Sheng Chen$^{1}$ and Yi He$^{4,*}$
\thanks{*This work was supported in part by the National Science Foundation (NSF) 
under Grants CNS-2245918, IIS-2245946, and IIS-2236578, and by the Commonwealth Cyber Initiative (CCI).}
\thanks{$^{1}$ Center for Advance Computer Studies, University of Louisiana, Lafayette, USA. 
        E-mails:        {\small \{leizhen.zhang1, sheng.chen\}@louisiana.edu}}%
\thanks{$^{2}$ Department of Computer Science, 
Old Dominion University, Virginia, USA. 
        E-mail: {\small l3li@odu.edu}}%
\thanks{$^{3}$ College of Computer and Information Science, Southwest University, China. 
        E-mail:        {\small wudi.cigit@gmail.com}}%
\thanks{$^{4}$ School of Data Science, 
William \& Mary, Williamsburg, Virginia, USA. 
        E-mail:        {\small yihe@wm.edu}}%
\thanks{* Corresponding Author: Dr. Yi He (yihe@wm.edu)}%
}

\title{ \Large \bfseries
Fairness-Aware Streaming Feature Selection with Causal Graphs
}

\begin{document}

\maketitle
\thispagestyle{empty}
\pagestyle{empty}

\begin{abstract}
This paper proposes a new online 
feature selection approach with an awareness of group fairness. 
Its crux lies in the optimization of a tradeoff 
between accuracy and fairness  
of resultant models on the selected feature subset.
The technical challenge of our setting is twofold:
1) streaming feature inputs, such that an informative feature may 
become obsolete or redundant for prediction if its information has been 
covered by other similar features that arrived prior to it,
and 2) non-associational feature correlation, such that 
bias may be leaked from those seemingly admissible, non-protected features.
To overcome this, we propose Streaming Feature Selection with Causal Fairness (\myAPP) that builds two causal graphs egocentric 
to prediction label and protected feature, respectively,
striving to model the complex correlation structure among 
streaming features, labels, and protected information.
As such, bias can be eradicated from predictive modeling by removing those features 
being causally correlated with the protected feature 
yet independent to the labels.  
We theorize that the originally redundant features for prediction 
can later become admissible, when the learning accuracy is compromised by 
the large number of removed features 
(non-protected but can be used to reconstruct bias information).
We benchmark \myAPP\ on five datasets widely used in streaming feature research,
and the results substantiate its performance superiority over six rival models 
in terms of efficiency and sparsity of feature selection and equalized odds
of the resultant predictive models.

\end{abstract}

\section{Introduction}

Streaming data deluges in online applications~\cite{hey2003data,slavakis2014modeling,lovelock2018forecast},
of which the huge volume tends to prohibit real-time decision-making 
that merely leverages manpower.
Decisions aided by computing models 
are hence increasingly becoming a norm,
which are about job or loan allocation~\cite{Dastin2018hiring,bogen2018help,binns2018s},
criminal recidivism~\cite{chouldechova2017fair},
fraud detection~\cite{bolton2002statistical},
online advertising and recommendation~\cite{evans2009online},
to name a few.
However, such algorithmic decisions 
may unfairly discriminate against certain social groups that are
constructed upon \emph{protected} user characteristics,
such as gender, age, and ethnicity~\cite{malleson2018equality}.

The pursuit of mitigating algorithmic biases originates from the research conducted 
by Friedman and Nissenbaum in 1996~\cite{friedman1996bias}
and has thrived within a decade~\cite{salimi2019interventional,mitchell2021algorithmic,hardt2016equality,kamishima2012fairness,
zafar2017fairness,jiang2020wasserstein,
friedler2014certifying,chiappa2019path,galhotra2022causal,orphanou2022mitigating},
mainly due to the recent popularity of 
data-driven prediction and classification models~\cite{binns2018s}.
In those models, 
bias exists in the form of spurious (or superficial) 
statistical correlation between feature vectors and target labels,
which represent users and decisions, respectively.
Thus, the key idea shared by existing studies lies 
in the prevention of spurious correlations in training data 
from influencing the resultant model.
Three main research thrusts stem from this idea~\cite{galhotra2022causal}: 
1) pre-processing methods that directly modify statistical distributions of 
training data ~\cite{calmon2017optimized, galhotra2022causal},
2) in-processing methods that impose fairness-based constraints 
on the objective function to regularize the training procedure~\cite{calders2010three,celis2019classification},
and 3) post-processing methods that scrutinize predictions and finetune 
the predicted label distributions to enforce outcome fairness with respect to prediction accuracy~\cite{kamiran2012data,fang2020achieving}.

Despite progresses,
these thrusts mostly overlook one prominent stage in the lifecycle 
of data-driven applications, i.e., \emph{data generation}.
Too often in data-driven businesses, 
practitioners can source and inquiry 
hundreds even thousands of features
for better user understanding and more precise user profiling~\cite{zhou2023febench}.
To wit, to detect fraudulent activities in online transactions, 
various features describing user behaviors are procured, 
such as the max/min/mean/top-k transactions
over the last 10 seconds, 3/24/48 hours, or 1/2/4 weeks, etc.
Consider the substantial loss of financial frauds~\cite{west2016intelligent}
and the fact that fraudsters can evolve their strategies to bypass the known indicative features~\cite{nicholls2021financial},
the banking industry urges to engineer more features and 
feature combinations to spotlight them.
We refer to this data generation continuum as 
\emph{streaming features}~\cite{wu2012online},
where the feature set is non-exhaustive
with new features constantly emerging.

%

\emph{How can algorithmic bias be eliminated from streaming features?}
This question remains open within the data mining community, 
and this study pioneers its first exploration.
Its main technical challenge is twofold.
On the one hand, the  bias structure between features and label may not be associational~\cite{salazar2021automated},
thereby complicating the strategy of predefining a subset of protected features
and then removing all new features entailing them in the streaming features.
On the other hand, new features may appear in high throughput, 
questioning the scalability of traditional feature selection methods being aware of fairness~\cite{jo2020lessons,schelter2019fairprep,galhotra2022causal}.
These methods operate and model the bias structure within a fixed feature space and thus must be reiterated each time a new feature is introduced.
They falter, if the processing time for each feature exceeds the 
timespan of feature generation.

%

To overcome the challenge, we propose a new algorithm, named 
\emph{Streaming Feature Selection with Causal Fairness} (\myAPP).
The key idea of \myAPP\ is to leverage a causal structure 
that enables non-associational modeling among protected features,
admissible features (not protected, but carry sensitive 
demographic information),
and the target labels.
Unlike previous causal fairness studies~\cite{salimi2019interventional,salazar2021automated}
that mostly postulate a known causal structure (which is next to impossible 
without domain knowledge), 
our \myAPP\ excels by learning causal structure from 
streaming features in an online manner.
Specifically, \myAPP\ dynamically constructs two causal graphs
being egocentric regarding the protected feature and label;
once a new feature arrives,
its topological positions on the two graphs 
are determined by its contribution to the algorithmic bias and prediction accuracy, 
respectively.
In particular, if this feature is irrelevant or redundant to the label,
it has no contribution to prediction and thus can be discarded.
On the contrary, 
if the feature is relevant as well as non-redundant to the protected feature,
it is \emph{inadmissible} by possibly leaking the bias of the protected user information 
through the causal structure, thus is also pruned.
In this study, we leverage d-separation~\cite{van2020finding}
to determine the feature positioning on causal graphs,
allowing for the gauging of feature relevancy and redundancy 
in terms of conditional independence, regarding the target variable, i.e., 
label or protected feature, in their respective causal graphs.

\smallskip\par\noindent
\textbf{Specific contributions of this paper are as follows:}
\begin{enumerate}
    \item Our work is the first study of online feature selection 
    in streaming features with respect to the algorithmic performance regarding both accuracy and fairness.

    \item We propose a new \myAPP\  algorithm to solve the problem, with its key idea lying in a dynamic modeling of causal structure among protected features, inadmissible and admissible features, and labels. 

    \item Extensive experiments on five benchmark datasets show 
    the superiority of \myAPP\, outperforming its six state-of-the-art competitors on average in the aspects of equalized odds, sparsity, and runtime by 52\%, 98\%, and 99\%, respectively, while maintaining an average accuracy that remains virtually unchanged.

    \item To champion reproducible research, our code and datasets are openly accessible at \url{https://github.com/zlzhan011/FSFS.git}.
\end{enumerate}


\section{Related Work}

\subsection{Fairness-Aware Machine Learning}
%
We can divide fairness-aware machine learning algorithms into three primary categories.
1) For pre-processing approaches~\cite {calders2009building, kamiran2012data, chiappa2019path, jiang2020wasserstein, feldman2015certifying, salimi2019interventional, galhotra2022causal, salazar2021automated}, 
they aim to directly modify statistical distributions of training data by assigning weights to features or labels to achieve a fair representation of protected and reference groups
with respect to their label assignments.
%
%
For example, \cite{calders2009building, kamiran2012data} propose massaging and reweighing techniques to adjust training data, ensuring predictions are independent of protected attributes.
\cite{chiappa2019path}  proposes a CGF framework that integrates causal graph structure learning with fairness regularization to minimize unfair edge weights, ensuring fair predictions by eliminating unfair causal effects.
\cite{jiang2020wasserstein} enforces statistical independence between model outputs and protected attributes, using Wasserstein-1 distances to align distributions across different groups.
\cite{salazar2021automated} 
resamples training data to reduce bias, ensuring predictions independent of sensitive attributes for causally fair prediction, 
which could be time-consuming due to the sampling process. 
%
\cite{salimi2019interventional}  firstly introduces the inadmissible features,
which are non-protected but could leak biased information from feature reconstruction.
To mitigate their impact, \cite{salimi2019interventional} proposes Capuchin
which repairs training data by implementing interventional fairness,
where both protected and  inadmissible features are d-separated 
in the causal structure. 
~\cite{galhotra2022causal} introduces a HypeR framework, which uses a conditional probabilistic causal model to handle what-if and how-to queries, reducing discrimination for fair causal predictions in databases.

2) For in-processing, the methods~\cite{zhang2019faht, aghaei2019learning} modify learning objectives or model 
structures to achieve fairness. For example, 
\cite{zhang2019faht} introduces a method 
integrating discrimination and information gain as a new 
splitting criterion when constructing a Hoeffding Tree. 
\cite{aghaei2019learning} proposes techniques incorporating 
regularization terms motivated by 
Mixed-integer programming to encourage low 
discrimination score. 
3) For post-processing, 
the methods~\cite{kamiran2010discrimination,hajian2015discrimination} 
adjust algorithm predictions to achieve fairness. 
They involve altering the confidence of classification rules or re-labeling predicted classes in decision trees to prevent biased decisions against protected groups. 
However, all these methods cannot be directly adapted to streaming feature selection,
mainly for two reasons.
First, they are offline and presume all features are accessible prior to the learning process,
which cannot be satisfied in our context wherein features emerge one at a time.
Second, they mostly postulate prior knowledge of domain data 
(e.g., a known causal structure) or 
model architecture (e.g., trees or tree ensembles),
thereby suffering from a generalization crisis when such assumptions do not hold in practice.
Our proposed approach excels in the sense that we allow a sequential feature input and 
we do not rely on any domain knowledge nor assuming the structure of classifiers,
thus is more general.


\subsection{Online Streaming Feature Selection}

The main task of the methods in this category~\cite{zhou2005streaming, zhou2006streamwise, wu2010online, you2021online, zhou2019online} is
to generate a feature stream and select the optimal feature subset, aiming to maximize prediction performance while keeping selected features as few as possible.
For example, \cite{zhou2005streaming, zhou2006streamwise} introduce an Alpha-investing method, which incorporates a new feature into the framework hinges on the p-value, with the impact of this addition evaluated through linear regression. 
To guide the selection of potential features, their technique relies on a heuristic understanding of the feature space's structure. 
However, this strategy may struggle with processing unmodified streaming features directly due to the practical challenge of acquiring comprehensive prior insights into such a structure.
To relax this requirement,
OSFS~\cite{wu2010online} 
builds a Markov blanket for the target labels based on statistical independence, 
approximating the causal relationship between streaming features and labels.
OCFSSF~\cite{you2021online} 
extends OSFS to consider spouse features of the Markov blanket by 
analyzing both conditional and unconditional independence relationships. 
However, these models mainly focus on the accuracy of prediction models
and do not consider algorithmic fairness.
Thus, once the protected and inadmissible features are wrongly incorporated 
into the selected feature subset, 
the resulting model will likely cause discrimination against protected groups,
leading to algorithmic bias.
To fill this gap, our proposed approach joins two objectives 
for optimizing both accuracy and fairness performance in the feature selection process.
In particular, our approach allows for 
new features originally being redundant to the label into the selected feature subset,
so as to remedy the information loss incurred by singling out 
the protected and inadmissible features for classification,
thereby balancing the tradeoff between accuracy and fairness. 

\section{Preliminaries}
\label{sec:Preliminaries}

\subsection{Problem Statement}

Let a sequence of features 
be $\mathcal{X}=\{ X_{i} \mid i = 1, \ldots, D \} \in \mathbb{R}^{D \times N}$,
where $X_i$ denotes the feature that emerges 
at the round $i$, and $D$
signifies the length of the feature stream.
%
The column vector $Y := \{ 0, 1 \}^{N}$ denotes the ground-truth labels.
There are $N$ instances in total.
We have $S \notin \mathcal{X}$ 
the \emph{protected} feature 
(e.g., gender, age, or ethnicity).
The feature subset selected at round $i$ is denoted by $\mathcal{F}_i^*$, $| \mathcal{F}_i^* | \leq i$.

Let $C$ denote the classifier trained on the selected features,
our feature selection problem is constrained by empirical risk (ER) 
and group fairness measurement (GFM):
\begin{equation}
    \min_{i=1, \ldots,T} \mathbb{E} [Y \neq C(\mathcal{F}_i^*)] + \lambda 
    \|\mathcal{F}_i^*  \|_0,
        ~\textnormal{s.t.}~ GFM(\mathcal{F}_i^*) \leq \epsilon, 
\end{equation}
where the first term amounts for classification errors and the 
$\ell_0$-norm enforces the number of features in $\mathcal{F}_i^*$ to be minimized.
The second GFM constraint 
can be implemented with demographic parity (DP)~\cite{verma2018fairness},
equalized odds (EO)~\cite{hardt2016equality, alghamdi2022beyond},
or other fairness metrics based on the domain requirements.
In this paper, we employ EO, defined:
\begin{align}
    GFM(\mathcal{F}_i^*) = \max_{y= \{ 0,1\} } 
    \big\{ \vert &\operatorname{P}(C(\mathcal{F}_i^*) =1 \mid S=0, Y=y) \nonumber \\
    - &\operatorname{P}(C(\mathcal{F}_i^*) =1 \mid S=1, Y=y) \vert \big\}.
    \label{eq:EO}
\end{align}
The intuition behind \eqref{eq:EO} is to encourage the data points from the
protected ($S=1$) and reference ($S=0$) groups to be predicted into the same 
class, with their maximum  difference in opportunity 
controlled within a threshold $\epsilon$.

%



\subsection{Causal Graph}
Causal graphs are directed and acyclic (i.e., DAGs).
Let $G = (\mathcal{V}, \mathcal{E})$ denote a causal DAG,
where $\mathcal{V}$ is the set of nodes,
and $V_i \in \mathcal{V}$ represents the $i$-th node with the feature $X_i$.
A causal DAG is egocentric if it maps all connections from 
the perspective of a target node~\cite{ellis2008learning}.
In this study, we deem protect feature $S$ or label $Y$ as the target node, generally denoted as $T$.
Given two features $X_i$ and $X_j$,
an edge $E_{ij}$ indicates the causal relationship between them,
where $X_{j} \rightarrow X_{i}$ means that $X_i$ is the immediate descendant of $X_j$.
All features must be the descendants of $T$ in our causal DAGs due to egocentricity. 
We propose to construct the edges from a Bayesian perspective,
which entails a set of variable independence definitions, as follows.

\subsection{Bayesian Causal Relationship}

To reason the causal relationship among features, we first define
their independence and conditional independence.

\begin{definition}[Null-Conditional Independence]\label{def:ConditionalIndependence} For any $i \neq j$, $X_i,X_j \in \mathcal{X}$ are null-conditional independence, iif 
P($X_{i}$$X_{j}$) = P($X_{i}$)P($X_{j}$),
denoted as $X_{i}$$ \bot $$X_{j}$ $\vert$$\varnothing $.
\end{definition}

\begin{definition}[Conditional Independence~\cite{koller1996toward}] For any $i \neq n \neq m$,
$X_{n},X_{m} \in \mathcal{X}$ are 
    conditional independence iif P($X_{n}$$\vert$$X_{m}$,$X_{i}$) = P($X_{n}$$\vert$$X_{i}$), 
    denoted as $X_{n}$$ \bot $$X_{m}$$ \vert $$X_{i}$.
\end{definition}

The topological position of a feature $X_i$
on $G$ can be located by its relevancy and redundancy w.r.t. other features:


\begin{definition}[Strong Relevance~\cite{kohavi1997wrappers}]\label{def:strongRelevance} A feature $X_i$ is deemed strongly relevant to $T$, iif 
$\forall X_j \subseteq \text{Pow}(G)$, it holds that $T \not \perp X_i \vert X_j$, where $\text{Pow}(\cdot)$ denotes powerset.
      
\end{definition}

\begin{definition}[Redundance~\cite{yu2004efficient}]\label{def:Redundancy} A feature $X_m$ is redundant to $T$, if there exists $X_j$ from the powerset of 
strongly relevant features,
such that 
$\text{P}(X_m$, $X_j) \neq 0$ and $T \bot X_m \vert X_j$.
\end{definition}

\begin{definition}[Irrelevance~\cite{wu2010online}] A feature $X_i$ is irrelevant, 
if it is not strongly relevant nor redundant to $T$.
\end{definition}

\begin{definition}[D-Separated]: 
If Z blocks all paths between X and Y, then X and Y are D-Separated and thus independent given Z.
\end{definition}

\subsection{Technical Challenges and Our Thoughts}

The main challenge arises from the existence of inadmissible features 
(a.k.a. proxy features~\cite{pessach2022review}) that 
are not protected, but can leak or be used to reconstruct the bias 
information conveyed by the protected features.
Examples of such inadmissible features abound, such as 
zipcode, which is commonly included in user profiling and model training but 
has been evidenced to reveal the user's ethnicity, resulting 
in predictions discriminating against certain races~\cite{kallus2022assessing}.
On the causal graphs generally, we can define:
\begin{definition}[Inadmissible Feature] A feature $X_i$ is inadmissible, denoted as $\mathcal IA$, 
if $X_i$ is not irrelevant to $S$. 
\end{definition}

This general definition would result 
in a performance tradeoff between accuracy and fairness,
because \emph{almost all} features are null-conditional independent 
from others from the linear algebraic perspective when $N$ is large.
Thus, it is most likely that many features 
in $\mathcal{F}_i^*$ are relevant to label $Y$
but are inadmissible.
Pruning all such features would incur 
information loss, degrading classification performance of the trained model.
In addition, computing the relevancy and redundancy 
of each incoming feature $X_i$
would require to compute the power set of 
the feature subset  selected up to $i$,
which entails combinatorial complexity 
depending on the size of $\mathcal{F}_i^*$.
The process is time and computationally intensive. 




 

To overcome the challenge, we propose to establish 
causal graphs for a sensitive feature $S$ and 
label $Y$ independently.
To ease notation, we denote them as 
the sensitive causal graph $G_{S}$ 
and the label causal graph $G_{Y}$.
We cast the problem of searching for inadmissible features into finding the overlapped features 
between the two graphs
and replace them with a set of 
\emph{admissible} features, denoted 
by $\mathcal A$.
The features in $\mathcal A$ reside 
outside of $G_{S}$ and $G_{Y}$,
which were originally deemed as redundant to $Y$.
An inspiring observation is that 
they are also conditional independent or null-conditionally
independent with the protected feature $S$,
indicating their d-separation from $S$.
Hence, it is possible that a feature subset in 
$\mathcal A$ may become redundant to $S$ 
(thus do not leak bias) and relevant to $Y$ 
(thus can provide classification information),
after the removal of inadmissible features.
As a result, 
locating this subset in $\mathcal A$
and replacing it with inadmissible features
can be a viable solution to optimize the 
tradeoff between accuracy and algorithmic fairness.


\section{The \myAPP\ Approach} 

Our proposed \myAPP\ proceeds in two steps.
First, two causal graphs egocentric to $S$ and $Y$ are built independently 
using Markov blanket, to model the non-associational contribution of an arriving feature to prediction and algorithm bias. 
Second, we search for an admissible feature set to replace 
those inadmissible features 
causally related to classification and protected information,
so as to remedy the predictive power loss incurred by removing them.
%


\begin{figure}[htbp]
    \vspace{-10pt}
    \centering
    \includegraphics[scale=0.5]{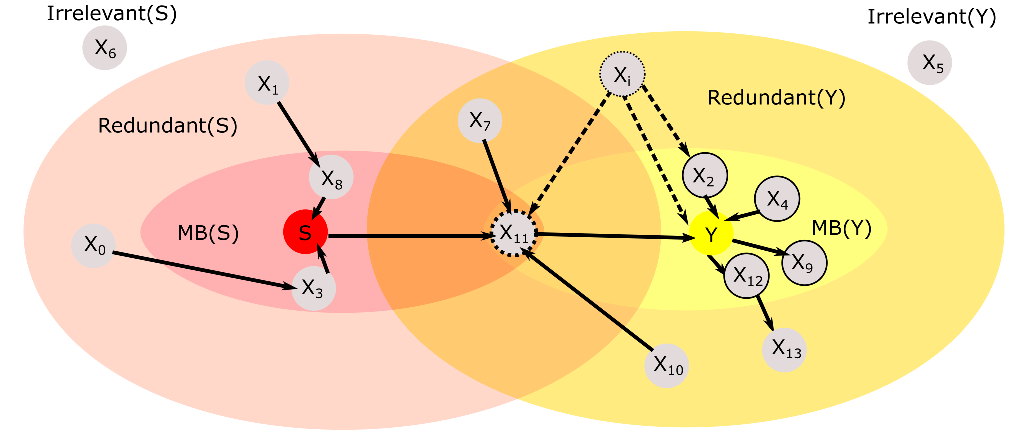}
    \vspace{-20pt}
    \caption{Causal graphs $G_S$ and $G_Y$. 
    1) Each small circle represents a feature or label with its name. 
    2) Varying shades of red areas represent $S$, $MB(S)$, $Redundant(S)$; 
    different shades of yellow areas represent $Y$, $MB(Y)$, $Redundant(Y)$; 
    white area represents $Irrelevant(S)$ and $Irrelevant(Y)$. 
    3) The dashed line represents the potential relationship.
    }
    \vspace{-5pt}
    \label{figure: $FS^2$ Structure}
\end{figure}

\vspace{-0.2cm}
\subsection{Causal Graph Construction with Markov Blanket}

To realize our idea, we use Markov blanket (MB) to build two causal graphs 
for sensitive feature $S$ and label $Y$ independently. 
The motivation for using MB is two-fold.
1) MB allows for an incremental modeling of feature correlation. In our context,
where new features emerge in sequence, and therefore,  most existing methods that require 
a complete feature space input falter.
2) MB causally d-separates an incoming feature from the target variable 
by retaining the strongly relevant features only,
which lifts the computational overhead required by powerset search 
for redundant features.

%


To do that,
we leverage the conditional independence tests including Fisher’s
z test and the $G_2$ test ~\cite{spirtes2000causation, neapolitan2004learning} (The details are explained in the supplementary materials). 
%
Specifically, given a target $T \in \{ Y, S\}$, 
for the $i$-th incoming feature $X_i$, 
we use conditional independence tests to identify three feature subsets and one corresponding relationship based on the MB for the $i$-th feature moment: $\mathcal StrongRelevant_{i}(T)$, $\mathcal Redundant_{i}(T)$, $\mathcal Irrelevant_{i}(T)$, and $COR_{MB}^{RE}$($i$)($T$). These are derived from the $(i-1)$-th feature moment, incorporating any changes that occur at the $i$-th feature moment.

The first step is to see whether $X_{i}$ and target $T$ satisfy the definition of null-conditional independence (Definition ~\ref{def:ConditionalIndependence}).
If $X_{i}$ satisfies, we will add $X_{i}$ into $Irrelevant_{i}(T)$; otherwise, we will add $X_{i}$ into the candidate features set, denoted as $\mathcal CFS(T)$.


The second step involves the redundancy analysis phase, which is triggered by the addition of a new feature to $\mathcal CFS(T)$. This phase filters the redundant features from $\mathcal CFS(T)$.
Once a new feature is added into $\mathcal CFS(T)$, we will do a loop for each feature $X_m$ in  $\mathcal CFS(T)$, to see whether there exist  $\exists X_j \subseteq \text{Pow}(\mathcal CFS(T) \setminus X_m )$ that satisfies $\text{P}(X_m$, $X_j) \neq 0$ and $T \bot X_m \vert X_j$ (Definition ~\ref{def:Redundancy}). 
If it exists, $X_m$ will be added to $\mathcal Redundant_{i}(T)$. Simultaneously, we record the corresponding relationship in $COR_{MB}^{RE}(i)(T)$, a dictionary where the keys are $X_j$ and the values are redundant feature $X_m$.
If it does not exist, we keep $X_m$ remaining in the $\mathcal CFS(Y)$.
After the loop, the features in $\mathcal CFS(T)$ excluding feature in $\mathcal Redundant_{i}(T)$ are $\mathcal StrongRelevant_{i}(T)$. 
Here, we observe that $\mathcal StrongRelevant_{i}(T)$ and $\mathcal MB_{i}(T)$
are equivalent based on Definition \ref{def:strongRelevance},
as both are the minimal feature subset for which $T \not \perp X_i \vert X_j$ holds for any other previously arrived features 
or feature combinations, i.e., $X_j \subseteq Pow ({X_1, \ldots, X_{i-1}})$.

\subsection{Optimizing the Accuracy-Fairness Tradeoff}

Based on the constructed causal graphs, at round $i$,
the inadmissible feature set $\mathcal IA_{i}$, admissible feature set $\mathcal A_{i}$ are: 
\vspace{-5pt}
\begin{equation} \label{eq:IA_formula}
    \mathcal IA_{i} = \mathcal MB_{i}(S) \cup  \mathcal Redundant_{i}(S)
\end{equation}
\begin{equation} \label{eq:IA_formula}
    \mathcal A_{i} = \mathcal MB_{i}(Y) \cup  \mathcal Redundant_{i}(Y) \setminus \mathcal IA_{i}
\end{equation}
\vspace{-18pt}

In figure \ref{figure: $FS^2$ Structure}, the inadmissible features set is the red area, and the admissible features set is the yellow area excluding the overlap area between red and yellow.

The intersection  $\mathcal MI_{i}$  between $\mathcal IA_{i}$ from $\mathcal MB_{i}(Y)$ is:
\vspace{-5pt}
\begin{equation} \label{eq:MI_formula}
    \mathcal MI_{i} = \mathcal MB_{i}(Y) \cup IA_{i}
\end{equation}
\vspace{-18pt}

Considering both accuracy and fairness, the initially selected feature set is $\mathcal MB_{i}(Y)$ removed intersection $\mathcal MI_{i}$:
\vspace{-5pt}
\begin{equation} \label{eq:RI_formula}
    \mathcal RI_{i} = \mathcal MB_{i}(Y) \setminus \mathcal MI_{i}
\end{equation}
\vspace{-18pt}

The motivation behind equation \ref{eq:RI_formula} is that
we can remove all inadmissible features from $\mathcal MB_{i}(Y)$, so that the selected features do not contain any sensitive information at all, but only contain accurate information.

If $\mathcal MI_{i}$ is $\varnothing$, it is an ideal situation. However, according to our observation, for most situations, the $\mathcal MI_{i}$ is not $\varnothing$.
If $\mathcal MI_{i}$ is not $\varnothing$, although removing it from $\mathcal MB_{i}(Y)$ can remove the sensitive information from selected features, it causes the accuracy information loss. In the figure ~\ref{figure: $FS^2$ Structure}, the $\mathcal MI_{i}$ contains feature $X_{11}$.

To compensate for the accuracy information loss caused by removing the $\mathcal MI_{i}$ from $\mathcal MB_{i}(Y)$, we can select partial features from $\mathcal Redundant_{i}(Y)$ to replace  $\mathcal MI_{i}$. 
Since some features are redundant given $\mathcal MI_{i}$ with target $Y$, if we remove $\mathcal MI_{i}$, they will be strong relevant with target $Y$.
For each feature in $\mathcal MI_{i}$, we can use the corresponding relationship $COR_{MB}^{RE}(i)(Y)$ to get the corresponding redundant features $\mathcal ICRF_{i}(Y)$  from $\mathcal Redundant_{i}(Y)$ and put them into $\mathcal ISF_{i}$. 

Because $\mathcal Redundant_{i}(Y)$ includes both admissible and inadmissible features, for the features $\mathcal AD1_{i}$ of $\mathcal ICRF_{i}(Y)$ belonging to the admissible set, incorporating $\mathcal AD1_{i}$ into $\mathcal ISF_{i}$ will enhance accuracy without compromising fairness. For example, in figure \ref{figure: $FS^2$ Structure}, $\mathcal AD1_{i}$ includes feature $X_{10}$.
\vspace{-5pt}
\begin{equation} \label{eq:AD1_formula}
    \mathcal AD1_{i} = \mathcal ICRF_{i}(Y) \cap A_{i}
\end{equation}
\vspace{-18pt}

Finally, for $\mathcal AD1_{i}$, the final selected feature set is $\mathcal{F}_i^*$:

\vspace{-18pt}
\begin{equation} \label{eq:FSF1_formula}
    \mathcal{F}_i^* = \mathcal RI_{i} \cup \mathcal AD2_{i} = \mathcal MB_{i}(Y) \setminus \mathcal MI_{i} \cup \mathcal AD1_{i}
\end{equation}
\vspace{-18pt}

However, for the other features $\mathcal AD2_{i}$ of $\mathcal ICRF_{i}(Y)$ in inadmissible features, considering $\mathcal MI_{i}$ belongs to either $\mathcal{MB}_{i}(S)$ or $\mathcal Redundant_{i}(S)$, $\mathcal AD2_{i}$, being a corresponding redundant feature linked to $\mathcal MI_{i}$, must be situated within $\mathcal Redundant_{i}(S)$. For example, in figure \ref{figure: $FS^2$ Structure}, $\mathcal AD2_{i}$ includes feature $X_{7}$. Our experiments indicate that $\mathcal AD2_{i}$
can increase the accuracy of information while the induced adverse impact on fairness can be accepted. 

\vspace{-10pt}
\begin{equation} \label{eq:AD2_formula}
    \mathcal AD2_{i} = \mathcal ICRF_{i}(Y) \cap \mathcal Redundant_{i}(S)
\end{equation}
\vspace{-18pt}

Finally, for $\mathcal AD2_{i}$, the final selected feature set is $\mathcal{F}_i^*$:

\vspace{-18pt}
\begin{equation} \label{eq:FSF2_formula}
    \mathcal{F}_i^* = \mathcal RI_{i} \cup \mathcal AD2_{i} = \mathcal MB_{i}(Y) \setminus \mathcal MI_{i} \cup \mathcal AD2_{i}
\end{equation}
\vspace{-18pt}

\vspace{-0.3cm}
\subsection{Time Complexity Analysis}
Our algorithms' main time consumption can be attributed to two key steps. Due to page limits, a detailed analysis of these steps is provided in the `C. Time complexity Analysis' section under `II. ALGORITHM' in the Supplementary.

\section{Experiments}

\vspace{-10pt}
\begin{table*}[!t]
	\centering
	\scriptsize 
	\caption{Results of experiments ACC$\pm$standard 
	deviation and EO$\pm$standard deviation 
	on 5 datasets using the LR classifier running 5 times. 
	1) The larger the ACC indicator, the better the model. 
	The smaller the EO indicator, the more fair the model.
	2) • denotes that our approach significantly 
	outperforms competitors according to paired t-tests 
	at the 95\% significance level.
					}

		    \begin{tabular}{
        >{\centering\arraybackslash}p{0.6cm}|
        >{\centering\arraybackslash}p{0.4cm}|
        >{\centering\arraybackslash}p{1.2cm}|
        >{\centering\arraybackslash}p{1.2cm}|
        >{\centering\arraybackslash}p{1.2cm}|
        >{\centering\arraybackslash}p{1.2cm}|
        >{\centering\arraybackslash}p{1.2cm}|
        >{\centering\arraybackslash}p{1.2cm}|
        >{\centering\arraybackslash}p{1.1cm}|
        >{\centering\arraybackslash}p{1.1cm}|
        >{\centering\arraybackslash}p{1.1cm}|
        >{\centering\arraybackslash}p{1.1cm}l
    }
			\hline
			\hline
				Dataset 												& Met 				& Baseline 				& Kamiran-massaging 	& Kamiran-reweighting  			&    Capuchin 			&     FairExp 			&        OSFS 			&    Remove S 			&        SFCF-RI 			&       SFCF-AD1 			&       SFCF-AD2 		 \\
			\hline																																
			\multirow{2}*{D1} 											& ACC				& .849$\pm$.002 		&       .845$\pm$.001 	&          .841$\pm$.005 		& .762$\pm$.020• 		& .755$\pm$.010• 		& .846$\pm$.003 		& .848$\pm$.002 		& .805$\pm$.004 		& .805$\pm$.004 		& .806$\pm$.004 		\\
																		& EO				& .115$\pm$.017• 		&       .227$\pm$.038• 	&          .079$\pm$.011• 		& .142$\pm$.013• 		& .025$\pm$.004 		& .101$\pm$.018• 		& .095$\pm$.020• 		& .035$\pm$.021 		& .035$\pm$.021 		& .036$\pm$.022 		\\	 
			\cline{1-12}	
			\multirow{2}*{D2} 											& ACC				& .679$\pm$.006 		&       .665$\pm$.015 	&          .658$\pm$.014 		& .624$\pm$.044 		& .551$\pm$.016 		& .682$\pm$.009 		& .680$\pm$.006 		& .550$\pm$.022 		& .550$\pm$.022 		& .576$\pm$.011 		\\
																		& EO				& .246$\pm$.017• 		&       .342$\pm$.041• 	&          .089$\pm$.032 		& .313$\pm$.075 		& .101$\pm$.025 		& .233$\pm$.022• 		& .236$\pm$.019• 		& .087$\pm$.026 		& .087$\pm$.026 		& .178$\pm$.036 		\\		 
			\cline{1-12}	
			\multirow{2}*{D3} 											& ACC				& .770$\pm$.025 		&       .711$\pm$.019 	&          .711$\pm$.019 		& .696$\pm$.041 		& .699$\pm$.031 		& .736$\pm$.032 		& .736$\pm$.032 		& .739$\pm$.021 		& .725$\pm$.012 		& .720$\pm$.012 		\\
																		& EO				& .287$\pm$.063• 		&       .229$\pm$.124 	&          .229$\pm$.124• 		& .287$\pm$.102• 		& .065$\pm$.049 		& .087$\pm$.032 		& .087$\pm$.032 		& .077$\pm$.039 		& .061$\pm$.033 		& .070$\pm$.044 		\\
			\cline{1-12}
			\multirow{2}*{D4} 											& ACC				& .806$\pm$.003 		&        .794$\pm$.010 	&          .793$\pm$.011 		& .796$\pm$.015 		& .651$\pm$.046• 		& .804$\pm$.004 		& .806$\pm$.002 		& .805$\pm$.002 		& .804$\pm$.003 		& .806$\pm$.002 		\\
																		& EO				& .044$\pm$.031 		&        .040$\pm$.022 	&          .041$\pm$.019 		& .065$\pm$.049 		& .188$\pm$.137• 		& .042$\pm$.023 		& .026$\pm$.019 		& .026$\pm$.019 		& .028$\pm$.015 		& .029$\pm$.028 		\\
			\cline{1-12}
			\multirow{2}*{\makecell{D5}} 								& ACC				& .856$\pm$.014 		&       .857$\pm$.012 	&          .827$\pm$.023 		& .770$\pm$.024• 		& .713$\pm$.000• 		& .848$\pm$.008 		& .855$\pm$.012 		&  .832$\pm$.020 		& .858$\pm$.017 		& .861$\pm$.012 		\\
																		& EO				& .421$\pm$.103• 		&       .427$\pm$.112 	&          .130$\pm$.059 		& .436$\pm$.107• 		& .280$\pm$.000 		& .500$\pm$.086• 		& .311$\pm$.074 		& .455$\pm$.134 		& .307$\pm$.083 		& .292$\pm$.069 		\\
			\cline{1-12}
			\multirow{2}*{\makecell{W/T/L\\-RI}} 						& ACC 				&       0/3/2 			&             0/3/2 	&                0/3/2 			&       2/3/0 			&       3/2/0 			&       0/3/2 			&       0/3/2 			&         --- 			&       0/4/1 			&       0/3/2 		\\
																		& EO 				&       3/2/0 			&             2/3/0 	&                2/2/1 			&       3/2/0 			&       0/5/0 			&       2/3/0 			&       2/3/0 			&         --- 			&       0/4/1 			&       1/2/2 		\\
			\cline{1-12}									
			\multirow{2}*{\makecell{W/T/L\\-AD1}} 						& ACC 				&       0/1/4 			&             0/3/2 	&                0/3/2 			&       2/3/0 			&       3/2/0 			&       0/3/2 			&       0/2/3 			&       1/4/0 			&         --- 			&       1/3/1 		\\
																		& EO 				&       4/1/0 			&             2/3/0 	&                1/3/1 			&       3/2/0 			&       0/5/0 			&       3/2/0 			&       2/3/0 			&       1/4/0 			&         --- 			&       1/4/0 		\\
			\cline{1-12}									
			\multirow{2}*{\makecell{W/T/L\\-AD2}} 						& ACC 				&       0/2/3 			&             0/3/2 	&                0/3/2 			&       2/3/0 			&       3/2/0 			&       0/3/2 			&       0/3/2 			&       2/3/0 			&       1/3/1 			&         --- 		\\
																		& EO 				&       4/1/0 			&             2/3/0 	&                1/2/2 			&       3/2/0 			&       0/4/1 			&       3/2/0 			&       2/3/0 			&       1/3/1 			&       0/4/1 			&         --- 		\\
			\hline
		\end{tabular}
\label{tab:Comparative result}
\end{table*}
\vspace{-10pt}

\begin{table*}[!t]
	\vspace{-5pt}
	\centering
	\small
	\caption{The average proportion of features selected, 
	based on 5 runs across 5 datasets. 1) D: Dataset;  2) \#: the number of features; 3) Met: Metrics name. 4) RSF: The ratio between selected features and the total number of features.. 5) NSF: the number of selected features;   
			}
	\vspace{-5pt}
	\begin{threeparttable}
		\begin{tabular}{
	    l|
	    >{\centering\arraybackslash}p{0.4cm}|
	    >{\centering\arraybackslash}p{0.5cm}|
	    >{\centering\arraybackslash}p{1.0cm}|
	    >{\centering\arraybackslash}p{1.2cm}|
	    >{\centering\arraybackslash}p{1.4cm}|
	    >{\centering\arraybackslash}p{1.1cm}|
	    >{\centering\arraybackslash}p{0.9cm}|
	    >{\centering\arraybackslash}p{0.6cm}|
	    >{\centering\arraybackslash}p{0.9cm}|
	    >{\centering\arraybackslash}p{0.8cm}|
	    >{\centering\arraybackslash}p{0.8cm}|
	    >{\centering\arraybackslash}p{0.8cm}
	}
		\hline
		\hline
		D    											&\#							  			&Met		&Baseline	 &Kamiran-massaging			&Kamiran-reweighting		&Capuchin			&FairExp					&OSFS     	&Remove S	   &SFCF-RI    				&SFCF-AD1			&SFCF-AD2 										\\
		\hline																																																																													
		\multirow{2}*{D1}								&\multirow{2}*{14}											&NSF			&14 		    &14 						&14 						&14 				&							&9			&13				&4							&5					&5													\\
		 													&														&RSF 	&1.0 			&1.0						&1.0						&1.0				&$>$1.0						&.642		&.928			&.285 						&.357 				&.357 												\\

		\cline{1-13}																																																																			
		\multirow{2}*{D2}								&\multirow{2}*{12}											&NSF			&12 			&12 						&12 						&12 				&							&5			&11				&1							&1					&2													\\
															&														&RSF 	&1.0 			&1.0						&1.0						&1.0				&$>$1.0						&.416		&.916			&.083  					&.083  			&.166 												\\

		\cline{1-13}																							                        			    						    						    								
		\multirow{2}*{D3}								&\multirow{2}*{25}											&NSF			&25 			&25 						&25 						&25 				&							&4 			&24				&3  						 &8     		&9 													\\
															&														&RSF 	&1.0 			&1.0						&1.0						&1.0				&$>$1.0						&.160		&.960			&.120 						&.320  			&.360 												\\

		\cline{1-13}																			                                        			    						    						    								
		\multirow{2}*{D4}								&\multirow{2}*{24}											&NSF			&24 			&24 						&24 						&24 				&							&8   		&23				&5 							&10    				&14													\\
															&														&RSF 	&1.0 			&1.0						&1.0						&1.0				&$>$1.0						&.333		&.958			&.208  					&.416     			&.583												\\

		\cline{1-13}										                                                                            			    						    						    								
																																																											
		\multirow{2}*{\makecell{D5}}					&\multirow{2}*{123}											&NSF			&123 			&123						&123						&123				&							&4			&122		 	&2							&14  				&104												\\
															&														&RSF 	&1 				&1 							&1 							&1 					&$>$1.0						&.032		&.991			&.016						&.113  			&.845												\\

	    \hline
	\end{tabular}

	\end{threeparttable}
	\label{tab:selected feature rate}
	
\end{table*}
\vspace{20pt}

\begin{figure*}[htbp]
	\vspace{-10pt}
	\centering
	\includegraphics[scale=0.34]{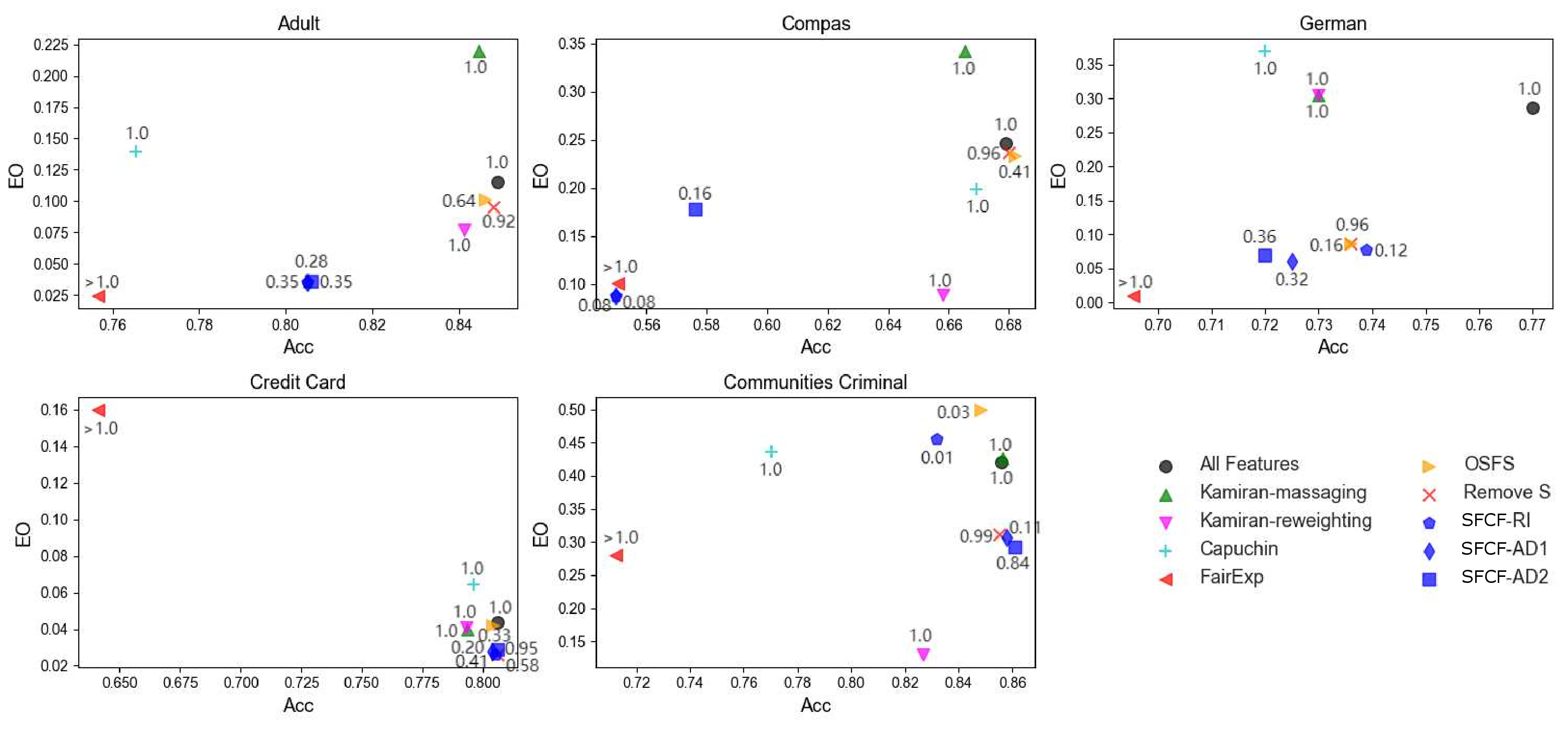}
	\vspace{-20pt}
	\caption{Experimental results, including ACC and EO, across 5 datasets using LR classifier. 
			 1) The higher the ACC indicator, 
			    the better the model performs. 
			 	Conversely, the smaller the EO indicator, 
				the fairer the model is considered. 
				Taking these two aspects into account, 
				a model that is closer to the bottom right corner 
				demonstrates better overall performance.
			 2) Our model, represented in blue, includes SFCF-RI, SFCF-AD1, and SFCF-AD2.
			 3) The numerical text surrounding the legend represents the average proportion of selected features. }
			 \vspace{-10pt}
			 \label{Comparisons with competitor(LR Model)}
			 \vspace{-3pt} 
\end{figure*}

\subsection{Experiment Setup}
\label{sec:Dataset}

In this work, we use five benchmark datasets, 
frequently employed in fairness literature, 
to robustly assess SFCF's effectiveness 
across diverse applications including income, 
credit card, and crime domains.
For the competitors, we have selected six distinct methods as competitors,
including Baseline(use all features, without features selection), Remove Sensitive, Kamiran-massaging/Kamiran-reweighting,
Capuchin, OSFS ~\cite{kamiran2012data, salimi2019interventional, salazar2021automated}.  
from various perspectives.
For the evaluation protocol, 
we leverage the Accuracy (ACC) and Equalized Odds (EO) 
to benchmark the experiments.
The detailed introduction of datasets, competitors, and evaluation protocol
was deferred into Section `A. Datasets', `B. Competitors', and `C. Evaluation Protocol' under `III. EXPERIMENTS' in the Supplementary due to the page limits.

\begin{table*}[!t]
	\centering
	\small
    \vspace{0.2cm}
	\caption{The average running time (in seconds) taken by different algorithms across five datasets over five runs. 
					}
	    \begin{tabular}{
        c|
        >{\centering\arraybackslash}p{1.0cm}|
        >{\centering\arraybackslash}p{1.2cm}|
        >{\centering\arraybackslash}p{1.3cm}|
        >{\centering\arraybackslash}p{1.2cm}|
        >{\centering\arraybackslash}p{1.4cm}|
        >{\centering\arraybackslash}p{1cm}|
        >{\centering\arraybackslash}p{1.3cm}|
        >{\centering\arraybackslash}p{1.2cm}|
        >{\centering\arraybackslash}p{1.2cm}|
        >{\centering\arraybackslash}p{1.2cm}
    }
		\hline
		\hline
		Dataset    			 						   			&Baseline				&Kamiran-massaging						&Kamiran-reweighting					&Capuchin					&FairExp									&OSFS   				&Remove S						&SFCF-RI    							&SFCF-AD1  				&SFCF-AD2   	\\
		\hline		                                                            			                        				                        				  	
		D1										       	    &4.925	    				&59.163									&59.782									&154.207 					&32882.201									&3.813  				&4.831							&3.471	  			    				&3.456	    				&3.779		 	\\
																																																
		\cline{1-11}		                                                				                        				                        				  				
		D2					 						   	    &3.326    					&69.354									&207.403								&207.462 					&3590.497									&2.273					&3.637							&1.681	  								&1.583	    				&1.685	     	\\
																																																
		\cline{1-11}						                                				                        				                        				      																													
					
		D3					 						   	    &5.417      				&28.342									&28.593									&66.220 					&3095.371									&1.062					&2.780							&1.147    								&1.068      				&1.130	     	\\
																																																
		\cline{1-11}						                                				                        				                        				      																															
		D4				 						    		&8.492     					&39.908									&41.176									&83.061 					&29326.990 									&5.339					&8.849							&4.341    								&4.957      				&4.291 	     	\\

		\cline{1-11}						                                                                                      																																								
         D5 & 3.441 & 42.447 & 40.409 & 175.345 & 658818.357 & 1.405 & 3.545 & 1.337 & 2.716 & 3.477 \\
		
	    \hline
	\end{tabular}
	\label{tab:Time}
	\vspace{-13pt}
\end{table*}
\vspace{-10pt}

\vspace{7pt}
\subsection{Experimental Results}
\label{sec:Results}

\textbf{Q1:} \textit{Do our algorithms outperform other methods?}

Three key observations can be drawn from Tables ~\ref{tab:Comparative result}, 
~\ref{tab:Time}, and ~\ref{tab:selected feature rate} to answer this
question.


First, our SFCF-AD1 algorithm does 
very well in the online streaming features scenario, 
giving the highest average EO score of 0.103, 
while keeping a competitive average accuracy of 0.748. 
Our algorithms, SFCF-RI, SFCF-AD1, and SFCF-AD2, 
are better than the Baseline algorithm by ratios of 39.82\%, 53.45\%, 
and 47.16\% in EO score, respectively, 
while only having a small decrease in accuracy ratios of 5.80\%, 
5.55\%, and 4.92\%. 

Second, compared with the competitors like Kamiran-massaging, 
Kamiran-reweighting, Capuchin, and FairExp, which 
use different mechanisms to address fairness 
via features or label processing, 
Our SFCF-AD1 beats all of them. 
Our SFCF-AD1 is better than them in the EO score by average 
ratios of 59.0\%, 8.8\%, 58.32\%, and 21.39\%, respectively. 
This confirms that conventional offline methods do 
sub-optimally in the online streaming feature scenario. 
Compared to OSFS, which has the same scenario with our algorithms, 
our algorithms show a significant decrease in EO score, 
dropping by ratios of 29.83\%, 46.21\%, and 37.17\%, 
while keeping a near-steady accuracy, 
which only dropped by ratios of 4.72\%, 4.44\%, and 3.75\%, respectively.

Last, as illustrated in Tables ~\ref{tab:Time} and 
~\ref{tab:selected feature rate}, 
our algorithms select the minimal 
average proportion of features (14.17\%). 
Compared to the Baseline, Kamiran-massaging, 
Kamiran-reweighting, Capuchin, and FairExp,  
our algorithms significantly reduce the proportion of the selected features 
by an average ratio of 85.83\%. 
 
It's worth noting that this selective use of features may result in some 
information loss, explaining the slight decrease in our algorithms' accuracy. 
However, this trade-off is acceptable given the significant reductions in 
EO score. Considering the aforementioned observations, 
we can state that our algorithm outperforms other methods.

\textbf{Q2:} \textit{How efficient do our algorithms excel 
in the context of streaming features?}

We can address this question 
through two observations. 
First, we compare OSFS, an online streaming feature selection method, to our proposed algorithmss: SFCF-AD1, SFCF-AD2, and SFCF-RI. 
In terms of EO score, 
our algorithms demonstrate an average decrease in the ratios of 
29.38\%, 46.20\%, and 37.17\% respectively, 
while the accuracy maintain stable, 
decreasing on average by the ratios of 4.72\%, 4.44\%, and 3.97\%. 
These results emphasize our algorithm's versatility 
in the online streaming feature scenario, 
striking a balance between accuracy and 
EO score.

Second, our algorithms show minimum time consumption, 
average 2.394 seconds. The average times reducing the time by ratios of 53.21\%, 46.17\%, 
and 43.90\% compared to the 
Baseline's average time. 
For offline methods, in terms of time consumption, 
our algorithms beat all of them. 
The time for SFCF-RI is only 5.01\%, 3.65\%, 2.09\%, and 0.01\% 
of Kamiran-massaging, Kamiran-reweighting, Capuchin, and FairExp respectively. 
Our algorithms are online and use incremental computation, 
meaning that when a new feature arrives, 
calculations are only done with the previously selected features, 
avoiding unnecessary computations. 
In contrast, offline methods need to compute each new feature 
with all the previously arrived features, 
involving lots of redundant calculations and wasting resources. 

\textbf{Q3:} \textit{How to optimize the tradeoff between 
accuracy-centric and fair-centric online feature selection?}

If we remove only the sensitive feature, 
it means that we want to retain more accurate information, 
and the selected features
are accuracy-centric. If we remove all the features in  $ G_{S} $, 
it means that we aim to achieve greater fairness, 
and the selected features are fair-centric.

First, for the accuracy-centric, 
we can answer the question by comparing 'Remove S' with 
Baseline and our algorithms. 
When compared to the Baseline, 
the accuracy of 'Remove S' remains almost unchanged,
with an average difference of just 0.88\%. 
However, the EO score shows an average decrease of 32.16\% across all 
five datasets. 
Notably, the decrease in EO score on the Credit Card dataset 
is a mere 4.06\%, whereas, on the other four datasets, 
the reductions are more considerable at 17.39\%, 26.12\%, 40.90\%, and 69.68\%, 
respectively. 
Now, comparing 'Remove S' with our algorithms, 
SFCF-RI, SFCF-AD1, and SFCF-AD2, 
our algorithms perform better, 
with EO score reductions by the ratios of 9.93\%, 31.39\%, 
and 19.86\%, respectively. 
These findings align with our expectations that 
removing only sensitive features can not tackle the fairness problem thoroughly,
since other features carry or contain sensitive information.

Second, for the fair-centric, we can answer it by comparing SFCF-RI with the 7 competitors. 
SFCF-RI is a method that eliminates all features 
in  $ G_{S} $. 
Here, SFCF-RI achieved a winning ratio of 57.14\% and a 
losing ratio of 2.86\% 
on the EO score when evaluated in a win/tie/loss format. 
In contrast, its performance on accuracy was less impressive, 
with a win ratio of only 14.28\% and a higher loss ratio of 28.56\%.
This suggests that while removing all features in  $ G_{S} $ can considerably improve the EO score, it doesn't fare as well in maintaining accuracy. 
This is due to the fact that SFCF-RI eliminates all features in  $ G_{S} $, some of which may contain label information beneficial for accuracy.

\textbf{Q4:} \textit{How well can the admissible features remedy 
the information loss incurred by the removal of features in  $ G_{S} $?}

First, comparing SFCF-AD1 and SFCF-AD2 with 
online and offline methods provide 
insight to answer this question. 
When comparing with online methods SFCF-RI, 
we observe that across all five datasets, 
SFCF-AD1 and SFCF-AD2 display a substantial decrease in EO score, 
reducing it by ratios of 23.82\% and 11.03\% respectively, 
while maintaining a stable average accuracy with a 
relative fluctuation of within 1\%. 
This suggests that $\mathcal AD1$ and $\mathcal AD2$ contain some information
of intersection, can remedy the information loss caused by removing the intersection.

Second, compared to the offline method Baseline, our algorithms decrease the accuracy by 
an average of 5.50\% and 4.82\% 
but significantly reduced the EO score by an average of 53.45\% and 45.64\%. 
Against offline methods Capuchin and FairExp, 
our methods outperform in both EO and accuracy, 
improving them by 2.57\%, 11.07\%, and 58.32\%, 21.39\%, respectively. 
In comparison to offline methods of Kamiran-massaging and Kamiran-reweighting, 
SFCF-AD1 shows superior performance in terms of EO score, 
averaging a decrease by ratios of 59.05\% and 8.80\%, respectively. 
In terms of accuracy, Kamiran-massaging and Kamiran-reweighting outperform 
SFCF-AD1, with higher ratios of 3.35\% and 2.29\%, respectively. 
Among these competitors, Kamiran-reweighting is closest to our performance,
but it utilizes 100\% of the features, our algorithms, 
SFCF-RI, SFCF-AD1, and SFCF-AD2, use only 14.17\%, 
25.56\%, and 45.97\% of the features respectively. 
This indicates that our algorithms achieve comparable effectiveness while using 
fewer features. 
Hence, the information loss incurred by the removal of features in 
graph $G_{S}$ can be remedied by the admissible features set.

\vspace{-0.3cm}
\section{Conclusion} 
Our research presents SFCF, an innovative method addressing the challenges of online feature selection in streaming data while ensuring fairness and maintaining predictive accuracy. By dynamically exploring the causal structure from incoming features in real-time, SFCF effectively captures the complex relationships among protected features, admissible features, and labels using the Markov blanket model, thereby mitigating algorithmic bias. Leveraging d-separation, we accurately gauge feature relevance and redundancy, enhancing fairness and accuracy. Our experiments show that SFCF outperforms state-of-the-art baselines across various metrics on benchmark datasets.


\bibliographystyle{IEEEtran}
\bibliography{ieee2024}

\end{document}